\pdfoutput=1
\def\year{2021}\relax
\documentclass[letterpaper]{article} 
\usepackage[switch]{lineno}
\usepackage{aaai21}  
\usepackage{times}  
\usepackage{helvet} 
\usepackage{courier}  
\usepackage[hyphens]{url}  
\usepackage{graphicx} 
\usepackage{graphicx}  
\usepackage{natbib}  
\usepackage{caption} 
\usepackage[percent]{overpic}

\usepackage{microtype}
\usepackage{tabularx}
\usepackage{graphicx}
\usepackage{subfigure}
\usepackage{booktabs} 
\usepackage{times}
\usepackage{epsfig}
\usepackage{graphicx}
\usepackage{amsmath}
\usepackage{amssymb}
\usepackage{multirow}
\usepackage{algorithm}
\usepackage{algorithmic}
\usepackage{subfigure}
\usepackage{amsmath}

\def\cput(#1,#2)#3{\put(#1,#2){\hbox to 0pt{\hss{#3}\hss}}}

\newtheorem{theorem}{Theorem}

\floatname{algorithm}{Experiment}
\title{THE DISTANCE BETWEEN THE WEIGHTS OF THE NEURAL NETWORK IS
MEANINGFUL\\}
\author{
    Liqun Yang \textsuperscript{\rm 1}
    Yijun Yang \textsuperscript{\rm 2}
    Yao Wang \textsuperscript{\rm 2}
    Zhenyu Yang \textsuperscript{\rm 2}
    Wei Zeng \textsuperscript{\rm 2}\\
}
\affiliations{
    \textsuperscript{\rm 1}Florida International University, School of Computing and Information Science\\
    \textsuperscript{\rm 2} Xi'an Jiaotong University, School of Computer Science and Technology\\
}

\begin{document}

\maketitle

\begin{abstract}
In the application of neural networks, we need to select a suitable model based on the problem complexity and the dataset scale. To analyze the network's capacity, quantifying the information learned by the network is necessary. This paper proves that the distance between the neural network weights in different training stages can be used to estimate the information accumulated by the network in the training process directly. The experiment results verify the utility of this method. An application of this method related to the label corruption is shown at the end.
\end{abstract}

\section{Introduction}
Since Dr. Hebb opened the door of machine learning in \cite{hebb1949organization}, people have obtained endless wealth from it. At the beginning of this century, neural networks' potential in machine learning tasks was discovered in many fields. With more and more people noticing this delicate model's power, applications based on the neural network develop rapidly and change the world gradually \cite{rumelhart1986learning,hochreiter1997long,fukushima1980neocognitron,lecun2015deep,hinton1994autoencoders,sajjadi2017enhancenet,he2017mask,an2015variational,arjovsky2017wasserstein}, which makes people eager to reveal the essence of the neural network.

As \cite{hornik1991approximation} proves, the network exists that is capable of arbitrarily accurate approximation to a specific function and its derivatives. Therefore, We need to find a suitable network structure for a specific task. To explain it, we denote one network's simulation capability as $I_0$, the information quantity of the relationship between two variables as $I_1$, and the information quantity of the training dataset as $I_2$. To get a trustworthy model, the basic requirement is $I_2 \geq I_0 >> I_1$. If $I_0<I1$, the model cannot simulate the relationship, which will make it hard to train to fine-tuned (underfitting). Else if $I_1 > I_2$, the dataset cannot express the relationship between the variables, which will mislead the model. Else if $I_0 > I_2$, there are too many options to simulate the relationship, which will make the model overfitted in most cases. In practice, $I_1$ is the exploration target, which can be estimated based on the background research, and $I_2$ can be calculated directly. Now, the question is reduced to \textbf{how to estimate the capability of one model.} There is no doubt that we can use the network's scale to estimate its capacity, but it is just a theoretical estimation. The theoretical upper limit of the human brain far exceeds what we can use. As a system with a similar structure, the network's real capacity is far less than its theoretical estimation. That is the reason why some small models can perform better than the bigger ones. What we take care of is the part we can use.

Therefore, the question is reduced to \textbf{how to quantifying the information learned by the neural network in training.} In \cite{tishby2000information}, the author uses the mutual information to analyze the changing of the information quantity in the neural network and put up with the concept called "information bottleneck." The information bottleneck theory describes the neural network's behavior and defines the optimal target, preserving the relevant information about another variable (maximize the bottleneck). One step forward, in \cite{tishby2015deep}, the author develops this method and puts it up with a tool, information plain, to visualize the neural network's behavior. Series of methods related to it are put forward \cite{alemi2016deep,higgins2016beta,yu2020understanding,achille2019information}, and people start to open the black box of the neural network.

\section{Motivation}
The similarity of the methods mentioned above is that they analyze the network's information by analyzing the data's changes. These methods have their rationality, and the weakness is also apparent.

Generally, errors in statistics cannot be eliminated. Comparing with the scale of the domain of the possible networks' input, the scale of the test case is too small, which will further magnify statistical errors. Specifically, let $I(X;\tilde{X})$ be the mutual information between the input data $X$ and the compressed representation (like the output of one layer) $\tilde{X}$.
\begin{equation}
    I(X ; \tilde{X})=\sum_{x \in X} \sum_{\dot{x} \in \tilde{X}} p(x, \tilde{x}) \log \left[\frac{p(\tilde{x} \mid x)}{p(\tilde{x})}\right]
    \label{eq_mutual_information}
\end{equation}
Based on Eq. \ref{eq_mutual_information}, we need to know the joint distribution $P(X,\tilde{X})$ and the distribution $P(\tilde{X})$, which two need to be counted in the experiment. Theoretically, for the neural network, $\tilde{X}$ is a continuous space. Discretization is necessary to count the probability distributions mentioned above. Different discretization functions will impact the result of observation significantly. Moreover, $\tilde{X}$ is tightly related to the layer's activate function based on the definition. If we just discretize $\tilde{X}$ evenly without further discussion, the activate function's feature will impact the observation result and mislead us. For example, in \cite{shwartz2017opening}, the author explains one behavior of the network. As mentioned in that paper, the experiment result indicates that the network's goal is to optimize the Information Bottleneck (IB) trade-off between compression and prediction, successively, for each layer. However, the related conclusion is challenged by other researchers. In \cite{saxe2019information}, the author proves that the two phases are just a special case caused by the non-linear activate function.

In a word, based on analyzing the relationship between input and output, the results will be impacted by the experiment's bias. Whereas directly analyzing the weights of the neural network can avoid the errors mentioned above.

In this paper, we will provide proof to show that the difference between the initial weights and the training's output weight can be used to estimate the quantity of information of the network accumulating in training. Moreover, we apply this method to analyze the impact of the label corruption. The corresponding experiment is shown at the end of this paper.

\section{The uncertainty of weights}
As Shannon mentioned in \cite{shannon1948mathematical}, information can be thought of as the resolution of uncertainty. Generally, we can use $H(X)$ to represent the chaos of the variable $X$ ($H(X)=\mathrm{E}[I(X)]$). The increase in information or energy will lead to a decrease in system entropy in the view of physics. Oppositely, if the entropy decrease is quantified, we can quantify the quantity of information accumulated by the system in this process.
\begin{equation}
    I = H_{P_0}(X) - H_p(X)
    \label{eq_information}
\end{equation}

Letting $I$ be the quantified information, which equal to the chaos reduction, we can use Eq. \ref{eq_information} to calculate it, where $H_{P_0}(X)$ and $H_{P}(X)$ is the system's entropy before and after receiving the information. To measure the information stored in the weights, we need to understand the uncertainty of the weights. Our viewpoint is the weights of the network are a variable in the training process. The appearance of a specific value of weights is uncertain because of the random factors in the training process, like the randomized initialization, optimizer (e.g., SGD), order of training data, etc. A certain training process is that in which there are no random factors. And it can be viewed as a special case of the uncertain ones.

There is another view to understanding the uncertainty of the weights. Generally, the calculation in the layer (including the activate function) is irreversible, which gives the capability of generalization to the network. Otherwise, for a network with specific weights, we can use the output to recover the input, which means the relationship between the input data $X$ and its corresponding compress representation $T$ is bijective, and the network degenerates to a codebook of input $X$. Therefore, for the external observer (only the network's input and output are visible), the network's weights cannot be calculated, which is the same as we cannot make sure the status of the Erwin Schrödinger's Cat \cite{marshall1997s}. The weights' value is hidden in an unknown wave function, like the cat's status is unknown after closing the box. Once we open the black box and observe the weights, the wave function collapses into a constant \cite{von2018mathematical}, which is the same as taking a sample from the current wave function.

In the view of information theory, the happening of an uncertain event (probability less than 100\%) provides the information to the receiver, called the variable's self-information \cite{jones1979elementary}. Respectively, we can receive the information through the appearance of specific weights. For a specific training stage, the appearance of a specific weight has a probability. With the network being trained continuously, the possibility distribution of the weights' appearance is changing respectively. Therefore, the information increase before and after the training can evaluate the information quantity provided by the training process.

\section{Probability space of neural networks weights}
For a specific neural network architecture, the status of one neural network can be identified by its weight uniquely. The set of all its possible weights is denoted as $\Omega$. $\mathbb{P}$ is the corresponding probability mass function. For any $\Sigma \in \mathbb{F}$, $p(\Sigma)$ is the corresponding probability. $(\Omega,\mathbb{F},\mathbb{P})$ is a probability space. The following discussion is based on this space. As we all know, the backpropagation (BP) algorithm, which is the basic method for network optimization, is a method working on Euclidean space. Therefore, essentially, this space is a Euclidean space and has two features.
\begin{enumerate}
    \item The dimension of elements in the space is high, which means there are too many elements in the space to enumerate.
    \item The possibility of one event happening $p(\{\omega\}) (\omega \in \Omega)$ is low, which means a single event occurs is almost impossible to observe.
\end{enumerate}
Moreover, two adjacent elements in this space might have different appearance probabilities in the specific training stage because of the limitation of computers' precision. For example, there are two weights $\omega_1$ and $\omega_2$, $\omega_1$ equals to $\omega_2 + \mathbb{\epsilon}$, where $\mathbb{\epsilon}$ is a vector whose components in each dimension are almost equal to the lowest precision error. For the same input, the output of these two layers might be different. This error will be magnified as the network depth increases, and the loss of these two weights can be different. The one with higher accuracy has a higher probability of appearing at the end of the training process. Therefore, the original space's discretization is hard to calculate directly, which means the prior probability $\mathbb{P}$ cannot be counted by the traditional method.

\subsection{The weights distribution visualization}
As Eq. \ref{eq_information} shows, to quantify the information accumulated by the network in the training process, we need to know the probability measure $\mathbb{P}_0$ and $\mathbb{P}$, where $\mathbb{P}_0$ and $\mathbb{P}$ are the probability mass function of the appearance of the weight before and after training respectively. The initialization of the weights is randomized, and $\mathbb{P}_0$ is an even distribution in a range defined by the initialization function. Now, the question is \textbf{how to estimate $\mathbb{P}$.}

\begin{figure}[htb]
    \centering
    \subfigure[] {
    \label{MDS_init}
    \includegraphics[width=0.22\textwidth]{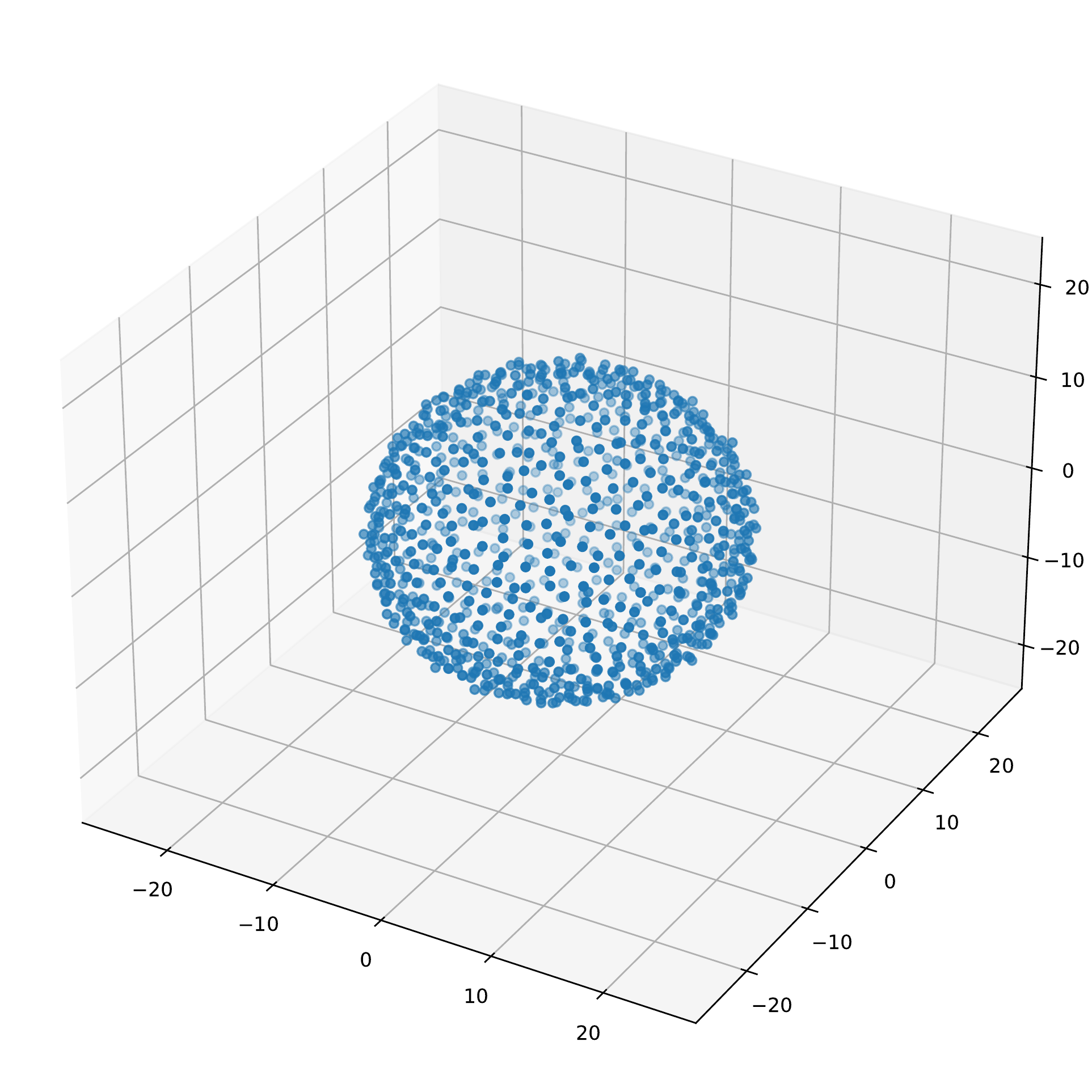}
    }
    \subfigure[] {
    \label{MDS_end}
    \includegraphics[width=0.22\textwidth]{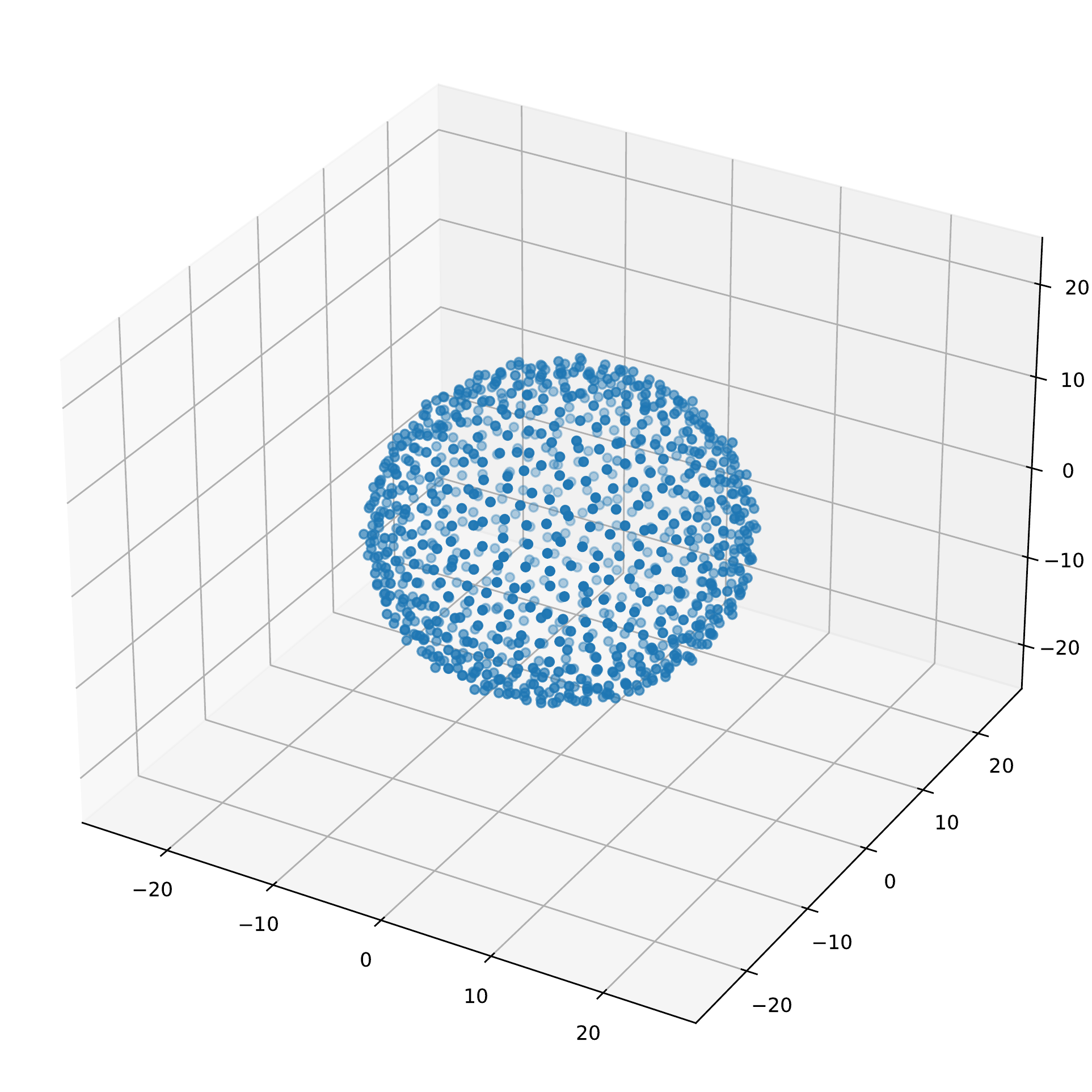}
    }
    \caption{The left shows the MDS result of the weights at the beginning. The right shows the MDS result of the weights at the end.}
    \label{MDS_fig}
\end{figure}

To observe the distribution of $\mathbb{P}$, we use the same training configuration to repeatedly train a specific network and collect the input weight (randomized) and output weights. Then, we use multiple dimensional scaling (MDS) to visualizing the level of similarity of individual weights. MDS is a method used to translate "information about the pairwise 'distances' among a set of n objects or individuals" into a configuration of $n$ points mapped into an abstract Cartesian space \cite{mead1992review}. The most important feature of MDS is that it can keep the Euclidean distance after dimensional reduction. The classic MDS algorithm \cite{wickelmaier2003introduction} is shown in Alg. \ref{MDS_fig}

\begin{algorithm}[htbp]
\caption*{\textbf{Algorithm 1} Multidimensional Scaling}
\begin{algorithmic}[1]
    \STATE Set up the squared proximity matrix $D^{(2)}=\left[d_{i j}^{2}\right]$, where $d_{ij}$ is the Euclidean distance between $i^{th}$ and $j^{th}$ elements.
    \STATE Apply double centering \cite{marden1996analyzing}: \\${\textstyle B=-{\frac {1}{2}}JD^{(2)}J}$ using the centering matrix ${\textstyle J=I-{\frac {1}{n}}11'}$, where ${\textstyle n}{\textstyle n}$ is the number of objects, $1$ being an $N ×1$ column vector of all ones.
    \STATE Determine the ${\textstyle m}$ largest eigenvalues ${\textstyle \lambda _{1},\lambda _{2},...,\lambda _{m}}$ and corresponding eigenvectors ${\textstyle e_{1},e_{2},...,e_{m}}$ of ${\textstyle B}$(where ${\textstyle m}$ is the number of dimensions desired for the output).
    \STATE Now, ${\textstyle X=E_{m}\Lambda _{m}^{1/2}}$, where ${\textstyle E_{m}}$ is the matrix of ${\textstyle m}$ eigenvectors and ${\textstyle \Lambda _{m}}$ is the diagonal matrix of ${\textstyle m}$ eigenvalues of ${\textstyle B}$.
\label{MDS_alg}
\end{algorithmic}
\end{algorithm}

Specifically, in this experiment, we use a TensorFlow CNN Demo \cite{TensorflowDemo} to identify the images in CIFAR-10 \cite{Krizhevsky09learningmultiple}. We fix all the super parameters in the training process and train the network from different scratches repeatedly. The experiment process is shown in Exp. \ref{Randomizedinit_exp} and the training configuration is shown in Tab. \ref{Tab_traininginfo}.

\begin{algorithm}
\caption{Input and output weights distribution by randomized training from scratches \label{Randomizedinit_exp}}
\begin{algorithmic}[1]
    \STATE Fix the training configuration, including all super parameters.
    \STATE Initialize 1000 networks randomly and save their initial value of weights.
    \STATE Train the networks to fine-tuned and save their value of weights at the end of the training.
    \STATE Use the MDS algorithm to visualize the input and output weights.
\end{algorithmic}
\end{algorithm}

As Fig. \ref{MDS_init} ($\Bar{r} = 1.28, std = 0.007$) and Fig \ref{MDS_end} ($\Bar{r} = 1.34, std = 0.12$) shows, all the points are distributed on the spherical surface evenly, which means their source vectors are also distributed evenly in the corresponding high-dimensional space.

As the reference, we add a constraint that limits the initial weights into a small range $\{w_0', w_0" \}$. Then train the network repeatedly. The experiment process is shown in Exp. \ref{Twoinit_exp}.
\begin{algorithm}
\caption{Input and output Weights distribution by randomized training from 2 specific scratches\label{Twoinit_exp}}
\begin{algorithmic}[1]
    \STATE Fix the training configuration, including all super parameters.
    \STATE Initialize 200 networks. Half of them use $w_0'$ as the initial weights, and the others use the $w_0''$.
    \STATE Train the networks to fine-tuned and save their value of weights at the end of the training.
    \STATE Use the MDS algorithm \ref{MDS_alg} to visualize the input and output weights.
\end{algorithmic}
\end{algorithm}

\begin{figure}[htb]
    \centering
     \includegraphics[width=0.22\textwidth]{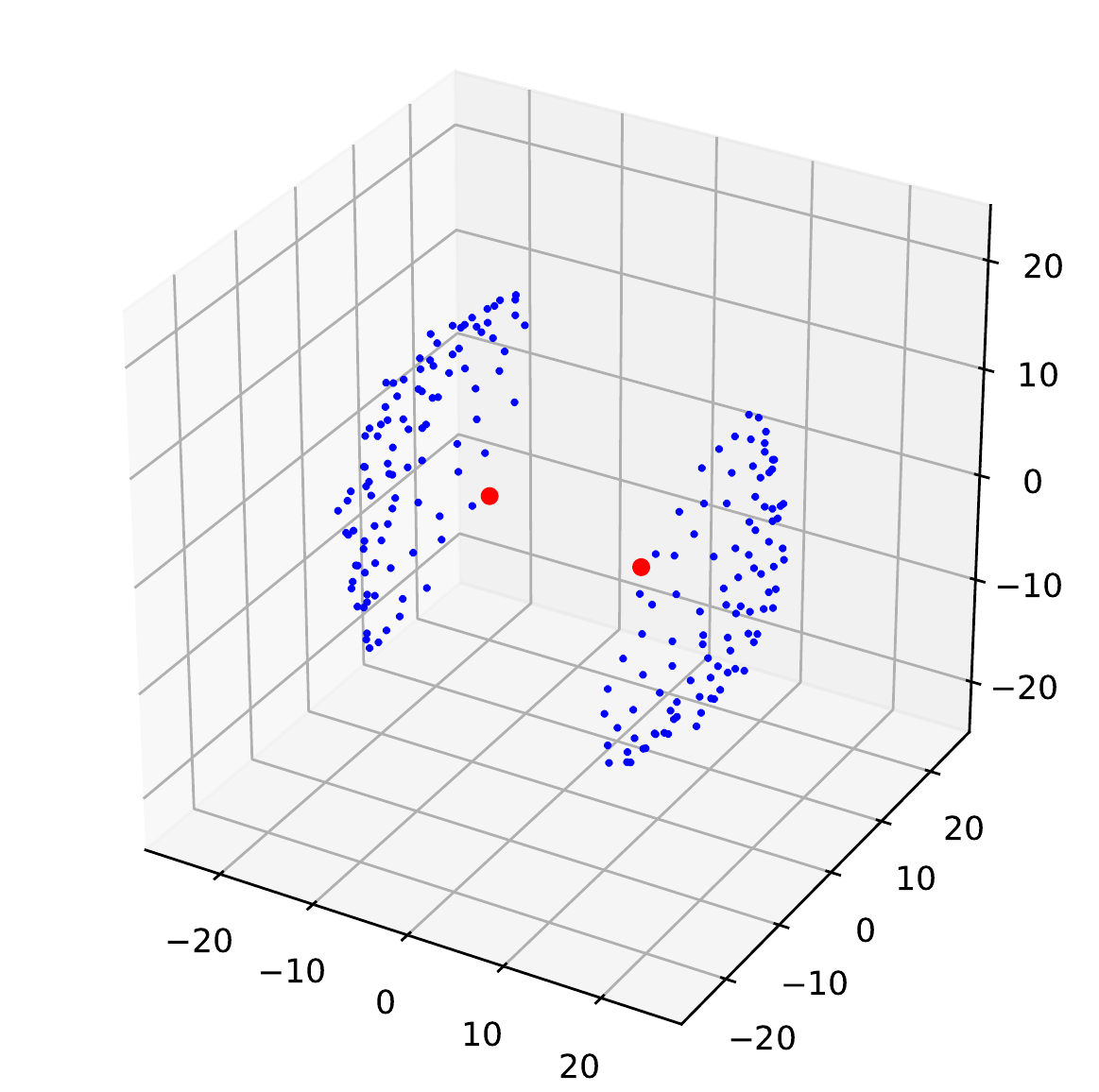}
    \caption{The initial weights are mapped to these two red points, and the output weights are mapped to these blue points.}
    \label{MDS_fig_polarized}
\end{figure}
We use the MDS algorithm to reduce the dimension of the initial weights and the output weights together. The output is shown in Fig. \ref{MDS_fig_polarized}. It shows that the output weights' mapping points (shown as the blue points) are distributed near their initial weights' mapping points (shown as the red points), which means the weights have higher appearance probability if its mapping point is in the region with more points. We can infer that if the mapping points distribute evenly in a region, their source's appearance probability is similar.

Based on the experiments mentioned above, we have Thm. \ref{even_distribution}
\begin{theorem}
    For a randomized training process, all the weights have the same appearance probability if they can appear.
    \label{even_distribution}
\end{theorem}

Letting $supp(p_0)$ and $supp(p)$ be the support of $p_0$ and $p$ respectively ($|supp(p_0)| \geq |supp(p_0)| > 0$), based on Thm. \ref{even_distribution}, we have
\begin{align*}
    H_{p_0}(X) &=  \log(\frac{1}{|supp(p_0)|}) \\
    H_{p}(X)  &=  \log(\frac{1}{|supp(p)|}). \\
\end{align*}
And we have
\begin{align*}
    I &= \log(\frac{1}{|supp(p_0)|}) - \log(\frac{1}{|supp(p)|}) \\
    &= \log (\frac{|supp(p)|}{|supp(p_0)|}).
\end{align*}

\begin{theorem}
\label{support_set_size}
    The information provided by training can be measured by the ratio between the support size before and after training.
\end{theorem}

Generally, if one training process cannot make the network converge into a stable status, the training fails. In this paper, we only discuss the successful training process ($supp(p) \subset Supp(p_0)$ and $I > 0$), and we have Thm. \ref{support_set_size}. Letting $r(p_0,p) = \frac{|supp(p)|}{|supp(p_0)|}$, the question is reduced to \textbf{how to estimate the ratio $r$ between the scale of support before and after the training.}

\section{Quasi-Monte Carlo method to estimate the set scale ratio}
Generally, the Monte Carlo method \cite{kroese2014monte} (MCM) can be used to estimate the scale shrink between one set and its subset. However, when one set's scale is much smaller than the other and elements are in a high dimensional space, the traditional MCM is not helpful. We provide a new quasi-Monte Carlo method (QMCM) to estimate the scale differences between two sets $ X $ and $X'$ when $X' \subset X$. Briefly, \textbf{the QMCM uses the expectation of the shortest distance to estimate the ratio between $X$ and $X'$.} For more details, we show the derivation below.

\subsection{Basic assumption}
\begin{figure}[htbp]
    \centering
    \includegraphics[width = 0.45 \textwidth]{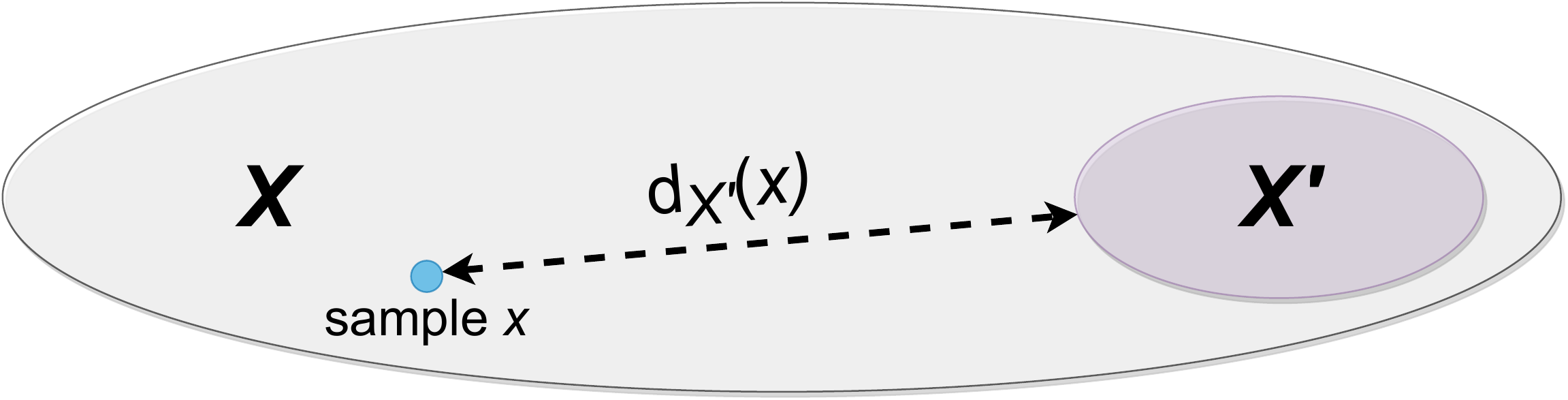}
    \caption{$d_{X'}(x)$ is the distance between $x$ and $X'$}
    \label{dm_definition}
\end{figure}

For a set $X$ which can be embedded into a measurable space and its non-empty proper subset $X'$, we define $d_{X'}(x)$ as the distance between $x$ and its closest element in $X'$ as Fig. \ref{dm_definition} shows, and we have $d_{X'}(x) = 0$ if $x \in X'$. Letting $D_{X'}(X)$ denote the sum of shortest distance for $x \in X$, we have
\begin{align*}
    D_{X'}(X) &= \sum_{x\in X}d_{X'}(x). \\
    \Bar{d}_{X'}(X) &= \frac{D_{X'}(X)}{|X|}(x \in X)
\end{align*}
Abbreviating $\Bar{d}_{X'}(X) $ as $\Bar{d}_{X'}$, for specific set $X$, $|X|$ is a constant.
If we want to use $\Bar{d}_{X'}$ to estimate $r$, we need to prove Thm. \ref{estimate_distance}.

\begin{theorem}
    Letting $r(X') = \frac{|X|}{|X'|}$, $\Bar{d}_{X'}$ is a monotonically increasing function of $r$ on for any subset $X'$ of $X$.
    \label{estimate_distance}
\end{theorem}

\textbf{Proof.} Letting $X''$ be the union of set $X'$ and the $\{x''\}$ ($x''\in X-X'$), we have
\begin{equation*}
    \frac{|X|}{|X''|}  < \frac{|X|}{|X'|}.
\end{equation*}
$d_{X'}(x'')$ is always positive, we have
\begin{equation*}
    D_{X'}(X) - d_{X'}(x'') < D_{X'}(X),
\end{equation*}
which leads to
\begin{equation*}
    D_{X'}(X-\{x''\}) < D(X,X').
\end{equation*}
Adding a new element $x''$ to $X'$ will update the distance from some elements to $X'$, noted as $\sum \Delta d$, and we have

\begin{equation*}
    D_{X''}(X) = D_{X'}(X-\{x''\})-\sum \Delta d.
\end{equation*}
Therefore, we have
\begin{equation*}
     D_{X'}(X-\{x''\}) - \sum \Delta d < D_{X'}(X-\{x''\}),
\end{equation*}
and we have
\begin{equation*}
    D_{X''}(X) < D_{X'}(X-\{x''\}),
\end{equation*}
which leads to
\begin{equation*}
    D_{X''}(X) < D_{X'}(X).
\end{equation*}
And we have
\begin{equation*}
    \frac{D_{X''}(X)}{|X|} < \frac{D_{X'}(X)}{|X|},
\end{equation*}

\begin{equation*}
    \Bar{d}_{X''} < \Bar{d}_{X'}.
\end{equation*}
And we have
\begin{equation*}
    r(X'') < r(X') \Rightarrow \Bar{d}_{X''} < \Bar{d}_{X'}.
\end{equation*}
The proof of other side is similar, and we have
\begin{equation*}
    r(X'') < r(X') \Leftrightarrow  \Bar{d}_{X''} < \Bar{d}_{X'} \square
\end{equation*}

Denoting the support of function $d_{X'}$ as $supp(d)$ and the expectation of $d_{X'}$ on $supp(d)$ as $\Bar{d}(supp(d))$. Based on the assumption,
\begin{align*}
    |supp(d)| = (1-r)|X|,\\
\end{align*}
we have
\begin{equation*}
    \Bar{d}(supp(d)) = (1-r)\Bar{d}(X)
\end{equation*}
For a specific $X'$, $r$ is a constant. When $1>>r$, we have
\begin{equation*}
    \Bar{d}(supp(d)) \approx \Bar{d}(X).
\end{equation*}
Now, the question is reduced to \textbf{how to estimate $\Bar{d}(supp(d))$.}

\subsection{The distribution of element-wised shortest distance}
Generally, we can use repeated random sampling to get a numerical approximation of $\Bar{d}(supp(d))$. If the cost of one sampling is high, we cannot take enough samples to prove the estimation accuracy. However, for a particular case, \textbf{when the mode of the distribution is equal to its mean, we can estimate the mean of the population with very few samples.}

To prove this, we need to analyze the distribution of $d_{X'}(x) (x \in supp(d))$. We use numerical simulation to analyze this function.
\begin{enumerate}
    \item Generate a set $X$ randomly with 10,000 elements, which are 100-dimensional normalized vectors.
    \item Select $r\%$ elements from $X$ randomly as the subset $X'$.
    \item Calculate the shortest distance from $x (x \in X-X')$ to $X'$ and count its distribution.
\end{enumerate}

\begin{figure}[p]
    \centering
    \includegraphics[width = 0.45\textwidth]{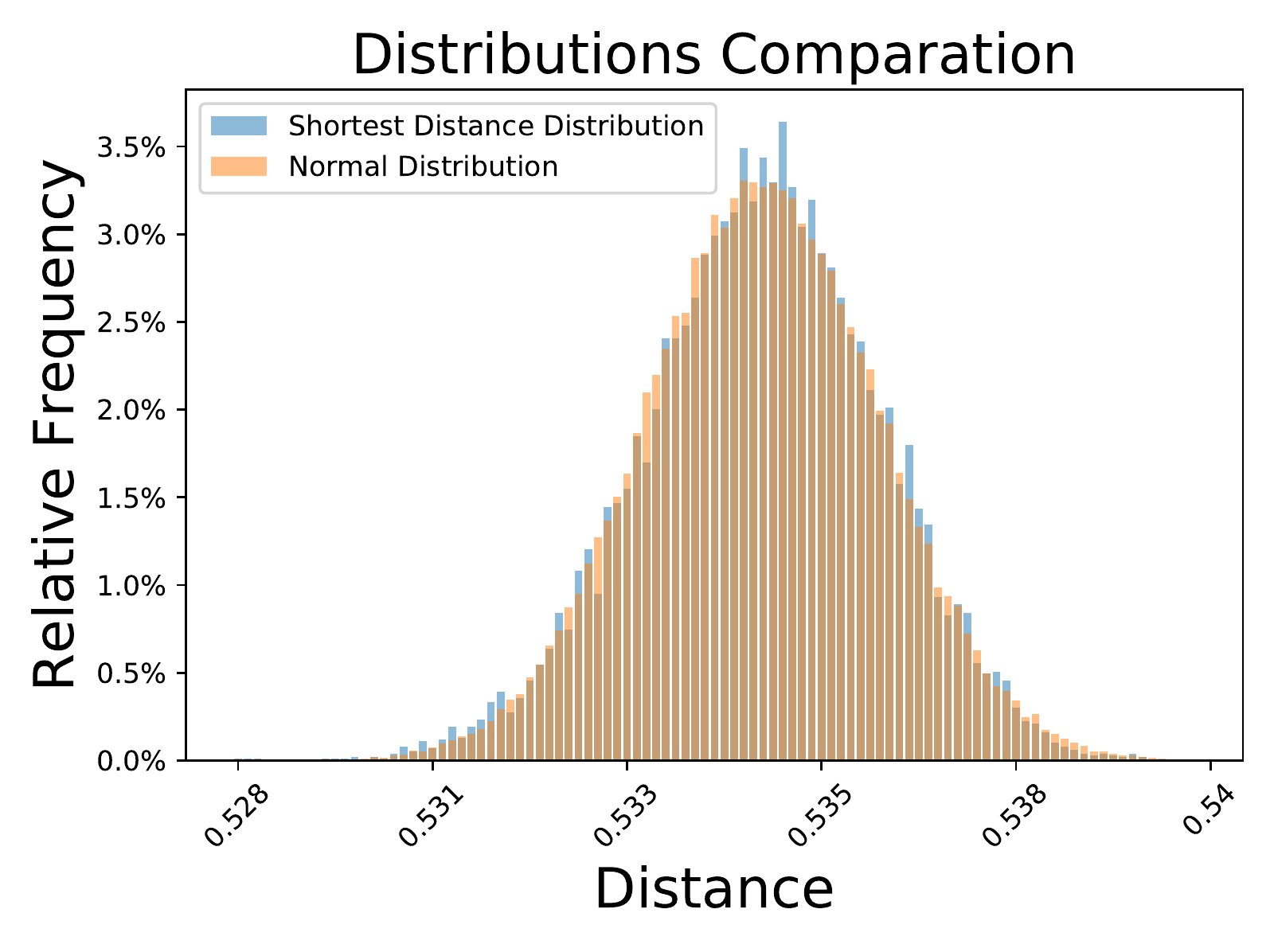}
    \caption{The shortest distance distribution is the distribution when $r = 10\%$, and the normal distribution is a randomized normal distribution with the same mean and standard difference value.}
    \label{Distribution_sample}
\end{figure}

\begin{figure}[p]
    \centering
    \includegraphics[width = 0.45\textwidth]{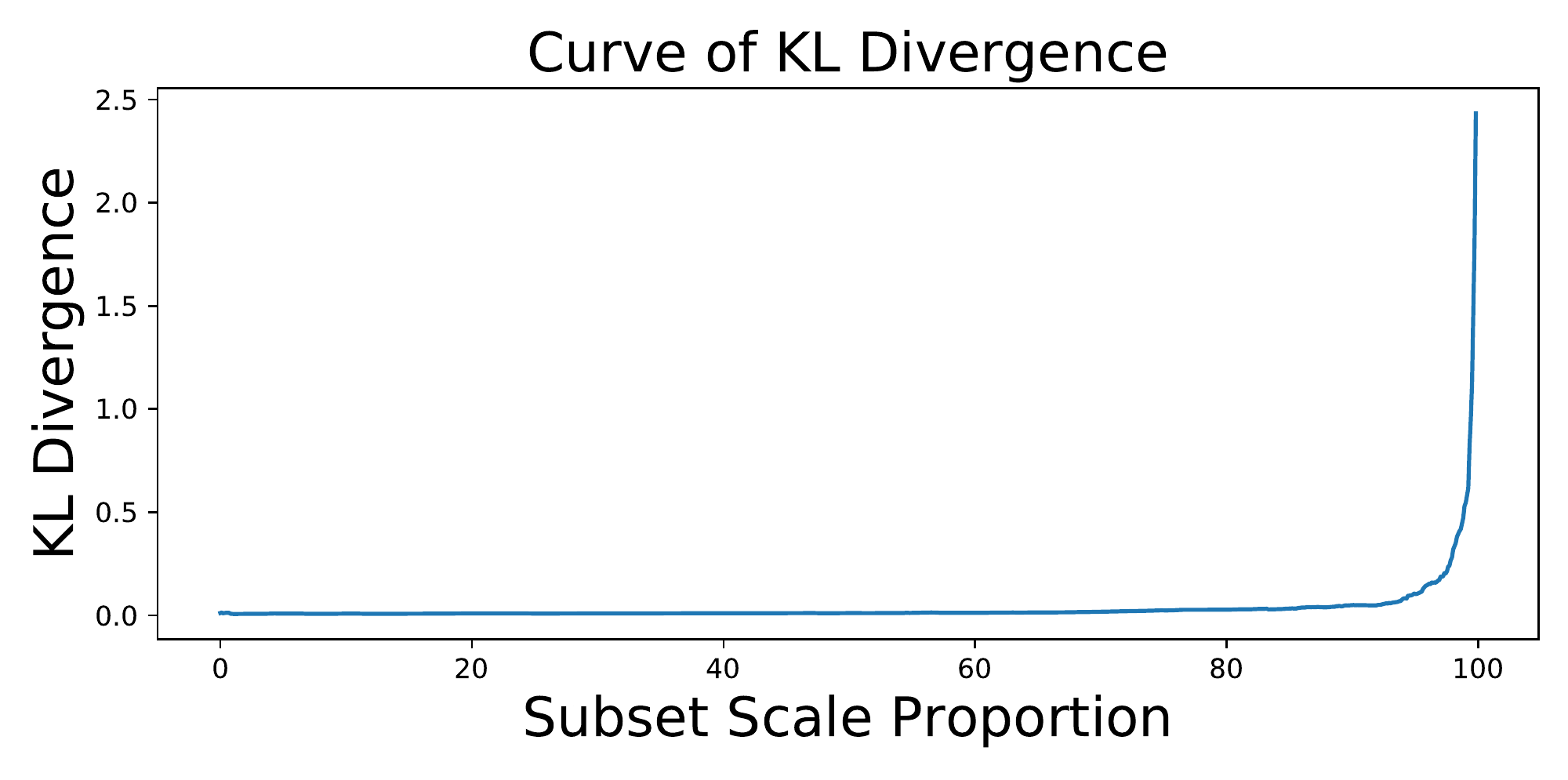}
    \caption{The curve of the KL divergence between the shortest distance distribution and the corresponding normal distribution.}
    \label{KL_curve}
\end{figure}

As an example, Fig. \ref{Distribution_sample} shows the result of the distribution when $r = 10\%$, and it is similar to the corresponding normal distribution (with the same mean value and standard difference value). Moreover, we change $r$ and calculate the KL divergence \cite{kullback1951information} between the shortest distance probability distribution ($P$) and the corresponding normal distribution ($Q$) as Eq. \ref{eq_KLdiv} shows. The result is shown in Fig. \ref{KL_curve}. Except for the cases when $r \geq 80\% $, the distribution trend is similar to a normal distribution, and we have Thm. \ref{support_distance_theorom}.

\begin{equation}
    D_{\mathrm{KL}}(P \| Q)=-\sum_{i} P(i) \ln \frac{Q(i)}{P(i)}
    \label{eq_KLdiv}
\end{equation}

\begin{theorem}
    For the elements in $supp(d)$, the distribution of their function value can be viewed as a normal distribution.
    \label{support_distance_theorom}
\end{theorem}

As a normal distribution, the mode of this distribution is equal to its mean. Letting $Y$ be the subset of $X$, when $c_v(d(y)) \leq t$ where $t$ is a threshold, we have
\begin{equation*}
    \Bar{d}(Y) \approx \Bar{d}(X)
\end{equation*}
And the QMCM method can be described as Alg. 2.

\begin{algorithm}[htbp]
\caption*{\textbf{Algorithm 2} QMCM for Subset scale estimation}
\begin{algorithmic}[1]
    \STATE Set the threshold $t$ and amount of samples $n$.
    \STATE Take $n$ samples from $X$ as $Y$ and calculate $Var(d(y))$.
    \WHILE {$c_v(d(y)) > t$}
    \STATE Remove the sample from $Y$ with the largest deviation from the mean.
    \STATE Resample and add the sample to $Y$.
    \ENDWHILE
    \STATE Output $\Bar{d}(Y)$.
\end{algorithmic}
\end{algorithm}
We can adjust the accuracy of the estimation by adjusting the value of $t$ and $n$. In the following experiment, we set $t = 0.3, n = 200$,

Unlike the traditional MCM, the novelty of this method uses the distance between related two points to estimate the ratio.

Abbreviating $\Bar{d}(Y)$ as $\hat{d}$, if we have two training process $t_1$ and $t_2$, we have
\begin{equation}
    \hat{d}_{t_1} < \hat{d}_{t_2} \Leftrightarrow I_{t_1} < I_{t_2}.
    \label{information_comparing}
\end{equation}
If $ \hat{d}_{t_1} < \hat{d}_{t_2}$, we have $I_{t_1} < I_{t_2}$. Oppositely, if $I_{t_1} < I_{t_2}$, we have $\hat{d}_{t_1} < \hat{d}_{t_2}$, which can be used to verify the correctness of our method.

\subsection{Apply on the neural network}
Correspondingly, for the network's training process, we have its initial weights and the output weights. The condition to implement QMCM to estimate the information is that the output weight is the closest one of the initial weights in $supp(P)$.

We select seven network models from simple to complex to verify this, TensorFlow MNIST classification Demo (classical version) \cite{TensorFlowMNIST}, TensorFlow CNN Demo \cite{TensorflowDemo}, GoogleNet \cite{szegedy2015going}, AlexNet \cite{krizhevsky2017imagenet}, ResNet \cite{he2016deep}, VGG \cite{simonyan2014very} and Yolo v3 \cite{redmon2018yolov3}. To ensure that these networks are used in scenarios that adapt to them, we select 4 dataset with different input scale and complexity, MNIST \cite{lecun1998gradient}, CIFAR-10 \cite{Krizhevsky09learningmultiple}, TensorFlow Flowers \cite{tfflowers} and Pascal VOC \cite{pascal-voc-2007}. We train each kind of model from scratch to fine-tune it with the same configuration 1000 times repeatedly. The basic information of the training is shown in Tab. \ref{Tab_traininginfo}. And then, we calculate the distance of arbitrary pairs of initial and end states. The result shows that all the output weight is the closest one of the initial weights in $supp(P)$. Therefore, we can use Eq. \ref{information_comparing} to compare the influence of the two training processes to the same model.

Moreover, we calculate the mean, standard difference, and coefficient of variation ($c_v$) value of the distance (see Tab. \ref{Tab_distancestatistic}). It shows that although the difference in the mean value among models is big, the coefficient of variation is still at a low level, which reflects the stability of this estimation.

\begin{table*}[]
    \centering
    \begin{tabularx}{\textwidth}{|c|X|X|X|X|X|X|X|X|X|X|}
    \hline
        \multicolumn{2}{|c|}{Model} & TF CNN & TF MNIST & GoogleNet & ResNet & VGG & AlexNet & Yolo v3\\
        \hline
        \multicolumn{2}{|c|}{Dataset} & CIFAR-10 & MNIST & TF Flower & TF Flower & TF Flower & TF Flower & Pascal\\
        \hline
        \multirow{2}{*}{Super} & LR & 0.01 & 0.01 & 0.01 & 0.01(decay) &0.01(decay) & 0.01(decay) & 0.01(decay)\\ \cline{2-9}

        \multirow{2}{*}{params}& Epoch & 10 & 10 & 20 & 40 & 20 & 20 & 20\\ \cline{2-9}
        & Batch Size & 128 &128 & 32 & 32&32 &32 &64 \\ \cline{2-9}
        & Optimizer & \multicolumn{7}{c|}{Stochastic Gradient Descent}\\
        \hline
         & Normlize & \multicolumn{7}{c|}{Yes} \\ \cline{2-9}
        Pre-& Mean Sub & \multicolumn{2}{c|}{No} &Yes &Yes &Yes &Yes  & No \\ \cline{2-9}
        process & Rescale &  \multicolumn{2}{c|}{Yes} &Yes &Yes &Yes &Yes & Yes\\\cline{2-9}
        & Standardize & \multicolumn{2}{c|}{Yes}&No &No &No &No  & Yes\\\cline{2-9}
    \hline
    \end{tabularx}
    \caption{Training configuration of experiments.}
    \label{Tab_traininginfo}
\end{table*}

\begin{table}[]
    \centering
    \begin{tabular}{|c|c|c|c|}
    \hline
    Network &  Mean & STD & $c_v$(\%) \\
    \hline
    TF CNN Demo & 4.026  & 0.053 & 1.316 \\
    TF MNIST Demo & 1.358 & 0.031 & 2.282 \\
    GoogleNet & 28.084 & 0.482 & 1.718\\
    ResNet & 8785.243 & 1172.836 & 13.350 \\
    VGG & 0.556 & 0.012 & 2.158 \\
    AlexNet & 4.849 & 0.589 & 12.163\\
    Yolo v3 & 26.863 & 6.286 & 23.400  \\
    \hline
    \end{tabular}
    \caption{The statistic result (mean value, standard difference and coefficient of variation) about the distance between initial and end states of neural networks.}
    \label{Tab_distancestatistic}
\end{table}

\section{Verification}
As mentioned in Eq. \ref{information_comparing}, we have  $I_{t_1} < I_{t_2} \Rightarrow \hat{d}_{t_1} < \hat{d}_{t_2}$, which can be used to verify the correctness of our method. We can construct two training process $t_1$ and $t_2$ such that $I(t_1) < I(t_2)$. Then calculate  $\hat{d}_{t_1}$ and $\hat{d}_{t_2}$ respectively.

Now, we need to construct these two training process $t_1$ and $t_2$. Based on the theory of information \cite{shannon1948mathematical}, higher entropy means more information. If we control all the other random factors and make $t_1$ contains more kinds of samples than $t_2$, we have $I(t_1)>I(t_2)$. Specifically, we use the same models to verify our method (see Tab. \ref{Tab_traininginfo}).

\begin{algorithm}[ht]
\caption{Training with different amount of labels\label{labels_exp}}
\begin{algorithmic}[1]
    \STATE Fix the training configuration, including all super parameters.
    \STATE Initialize 5 networks $\{m_i\} (i \in [0,4])$.
    \STATE Use data with $(20\times i)\%$ labels to train the network separately and record the weights changing.
    \STATE Calculate $d(w,w')$ of each weight, where $w$ is the current weights, $w'$ is the output weights of the same training process.
\end{algorithmic}
\end{algorithm}

In \cite{wang2020generalizing}, the author proves that few samples can also be used to guide the network to complete a complex task. Because of that, to ensure  $I(t_1)>I(t_2)$, we control the numbers of labels. There are $[20\%,40\%,60\%,80\%,100\%]$ kinds of samples are used in the training. To eliminate the impact of the weights update numbers, we train the same amount of steps in all the model training process. Finally, we calculate the $\hat{d_{t_i}}$ for each training process (see Exp. \ref{labels_exp}).
\begin{figure*}[p]
    \centering
    \subfigure[TF CNN Demo] {
    \includegraphics[width=0.45\textwidth]{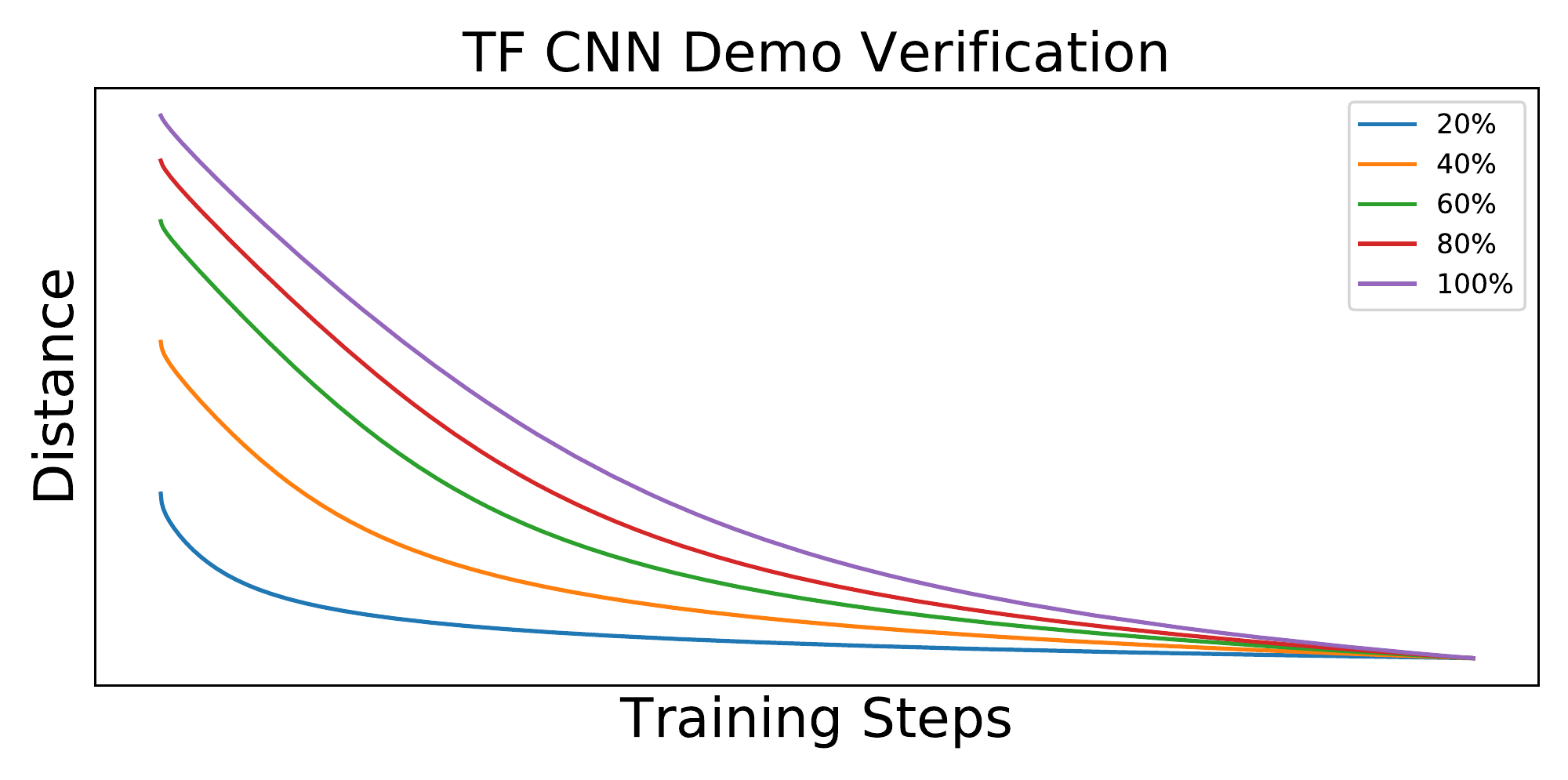}
    }
    \subfigure[TF MNIST Demo] {
    \includegraphics[width=0.45\textwidth]{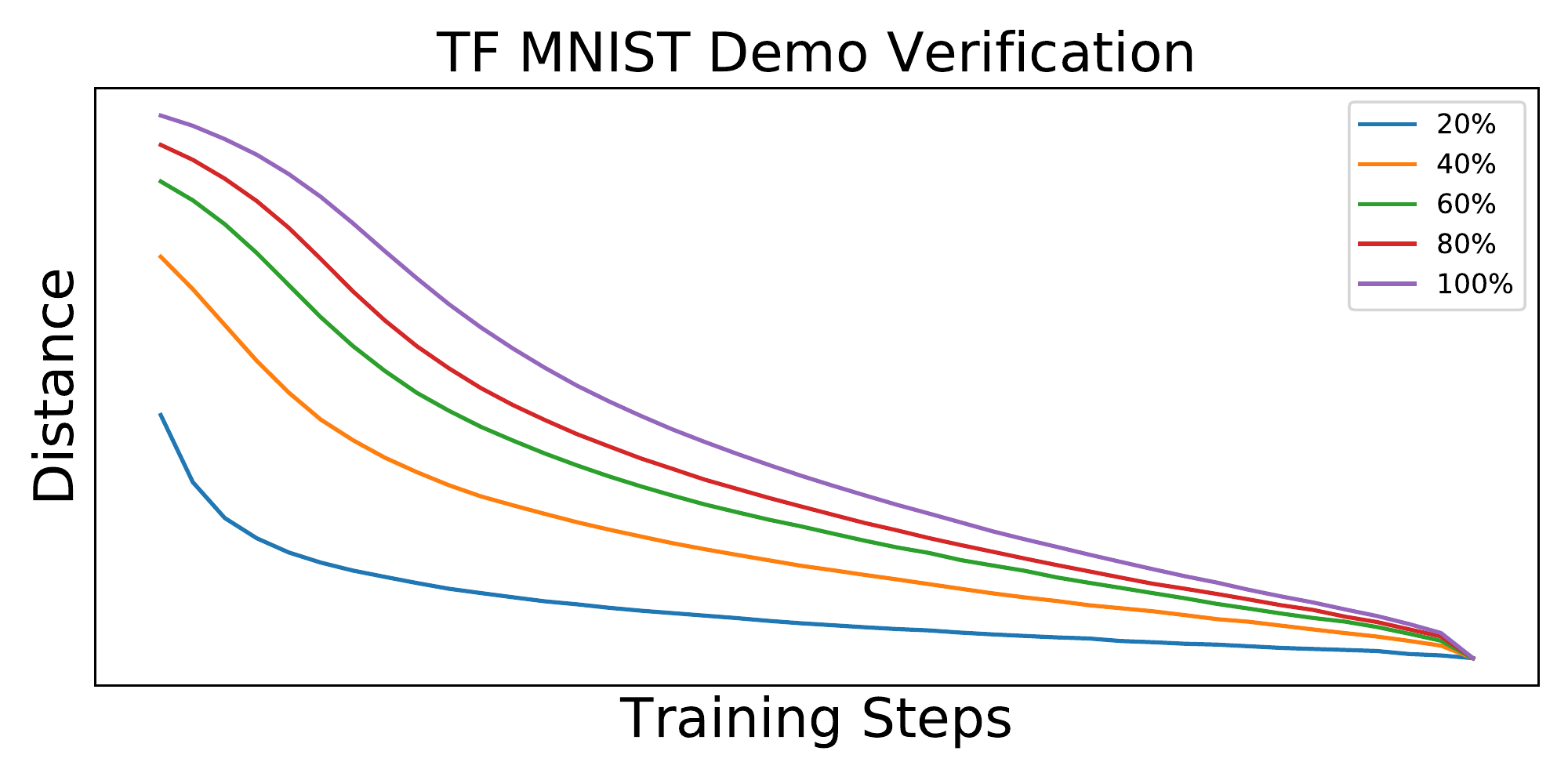}
    }
    \subfigure[GoogleNet] {
    \includegraphics[width=0.45\textwidth]{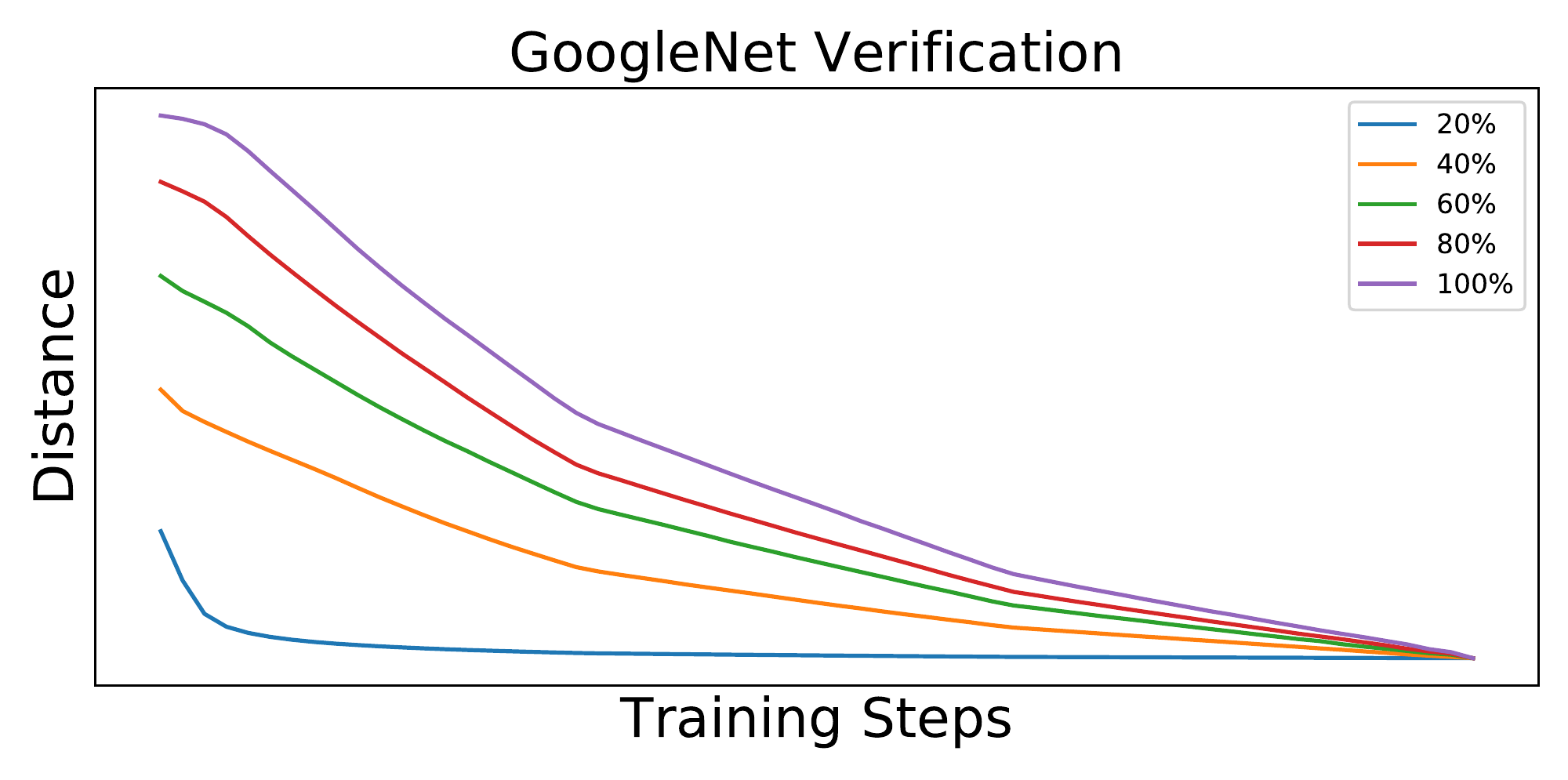}
    }
    \subfigure[VGG] {
    \includegraphics[width=0.45\textwidth]{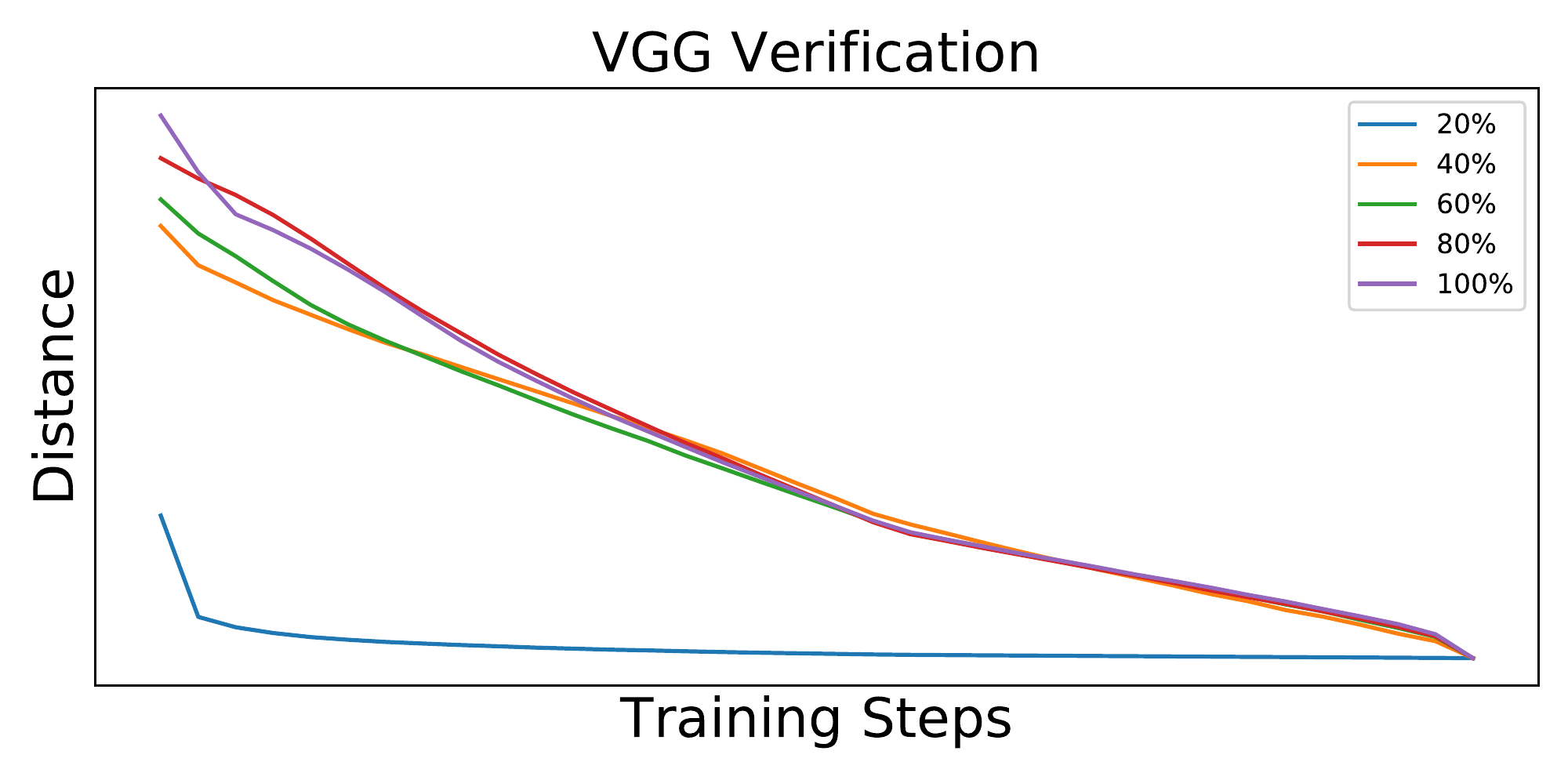}
    }
    \subfigure[ResNet] {
    \includegraphics[width=0.45\textwidth]{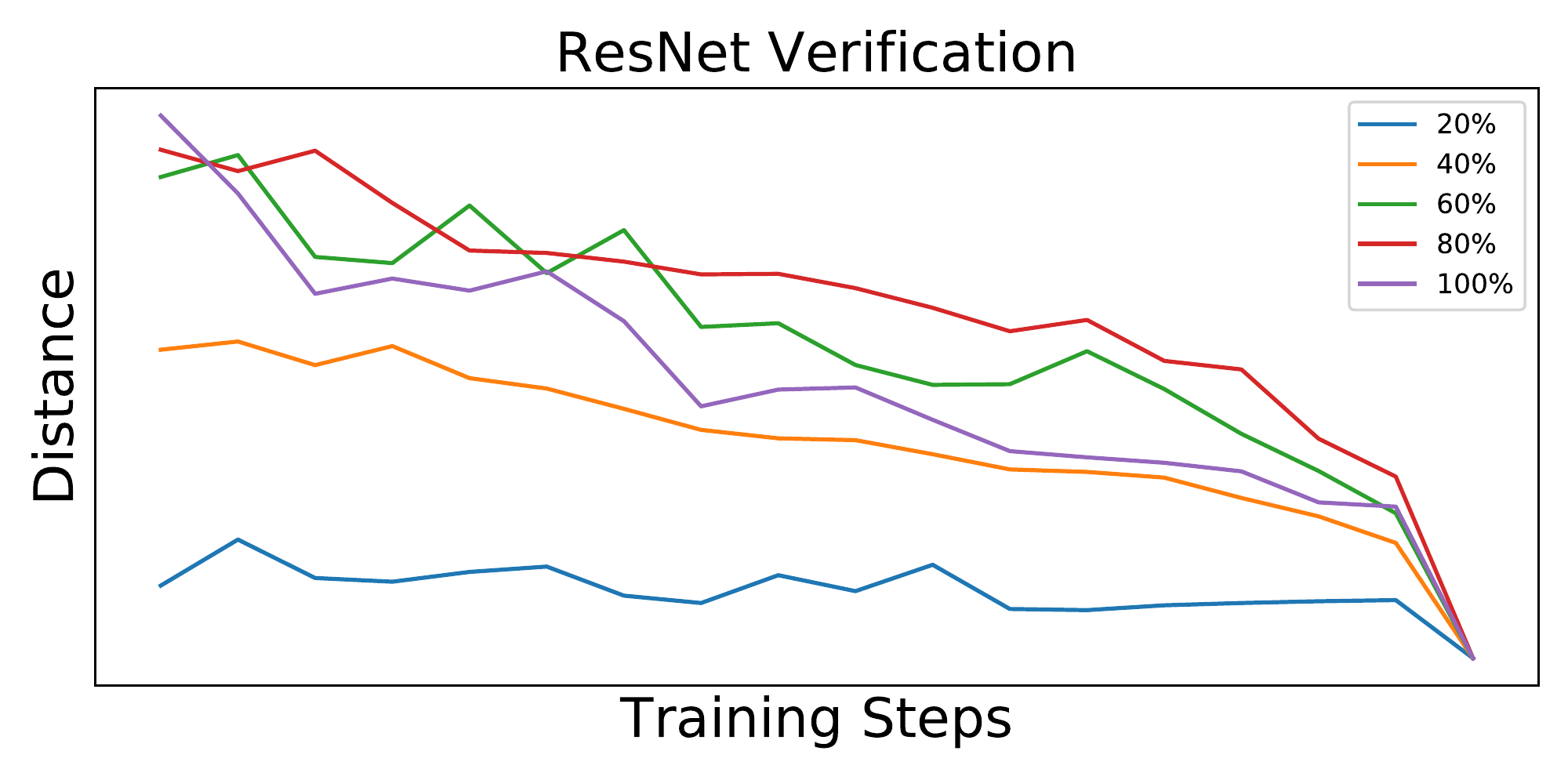}
    }
    \subfigure[AlexNet] {
    \includegraphics[width=0.45\textwidth]{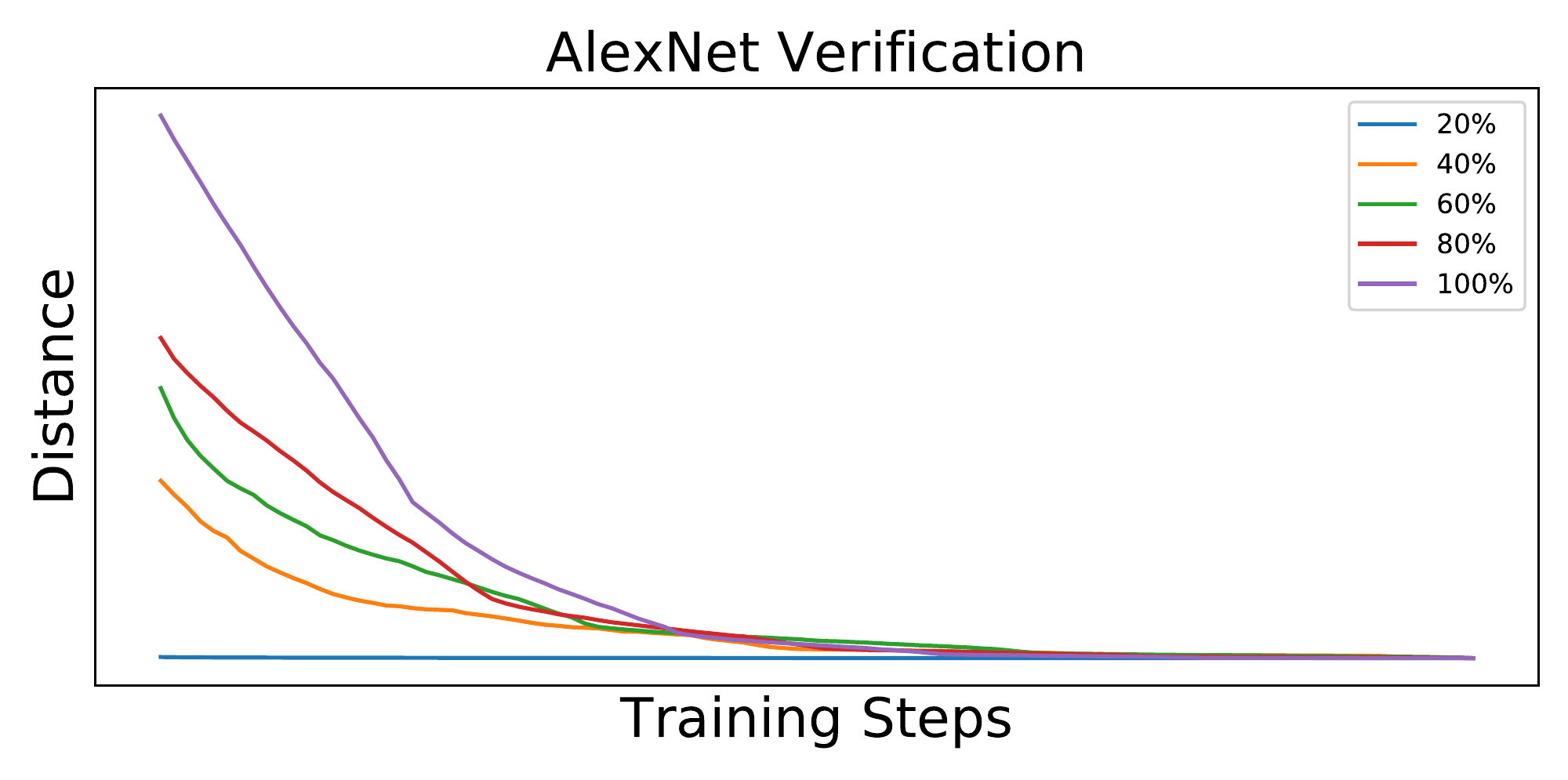}
    }
    \subfigure[Yolo] {
    \includegraphics[width=0.45\textwidth]{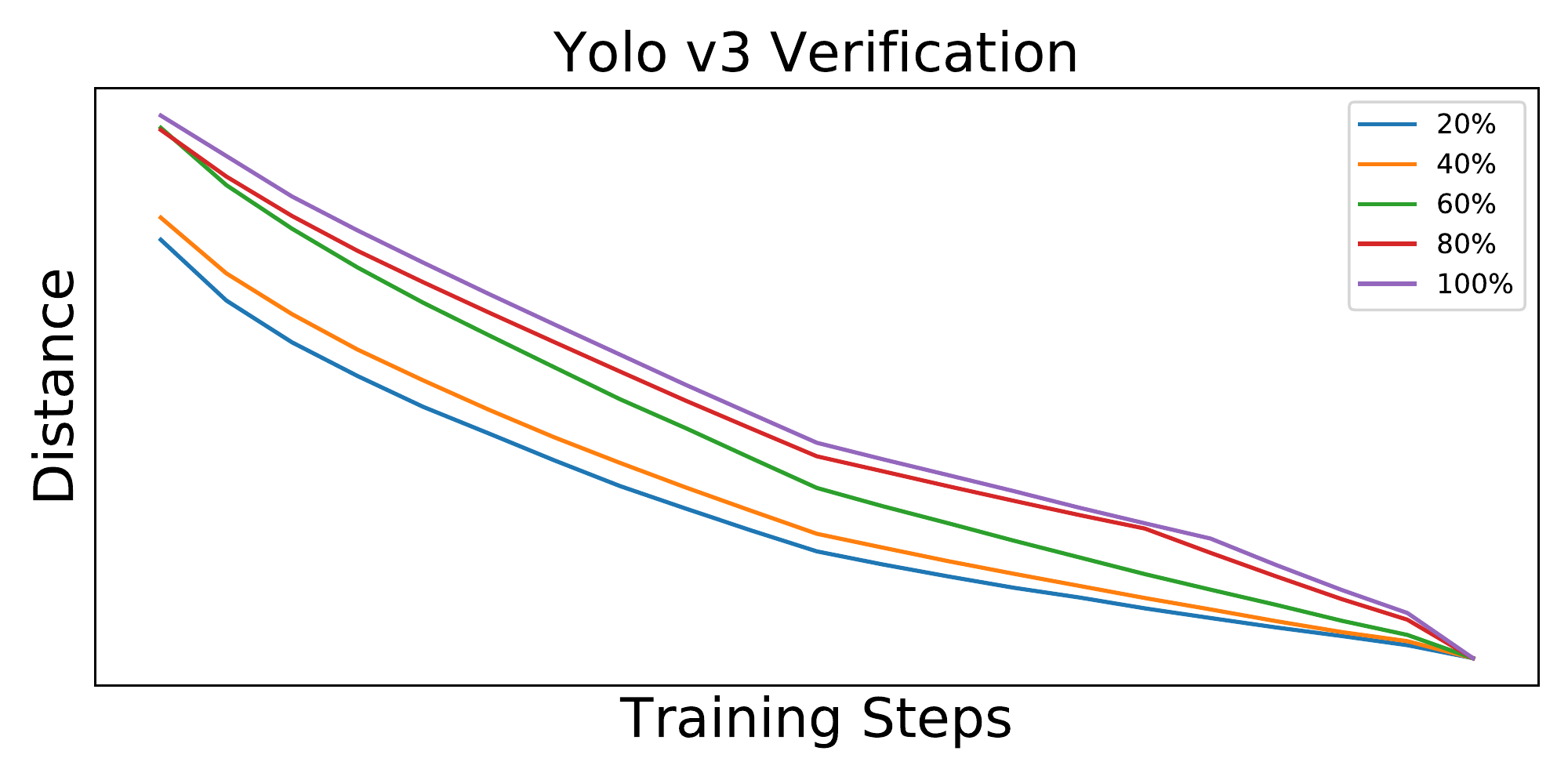}
    }
    \caption{The summary of the Exp. \ref{labels_exp} results of the verification.}
    \label{ed_1}
\end{figure*}

As Fig. \ref{ed_1} shows, in all experiments, as the $I$ increases, $\hat{d_{t_i}}$ increases, which verifies the utility of our method.

\section{Application: impact of label corruption}
The most significant application is to evaluate the training process.

In most cases, the trained network's performance is the only measure in the evaluation of the training process. However, as the author mentioned in \cite{geirhos2020shortcut}, some of the information in the dataset can be the "shortcut" to complete tasks, which leads to the training's failure. If the test dataset has similar "shortcuts," this failure will be unnoticeable without complex analysis. Moreover, the Clever Hans effect has been observed in the early version of BERT \cite{devlin2018bert} when it completes the argument reasoning comprehension task. The network makes the correct judgment based on a hidden trick but not the logic we want it to learn, which Niven and Kao noticed in \cite{niven2019probing} . By analyzing the quantity of information, we can reveal the essence of the networks' learning process and avoid being misled by the errors mentioned above.

Here, we use the impact of the label corruption as an example to show our method's application. For the impact of the label corruption, as the author mentions in \cite{zhang2016understanding}, some networks can build the relationship between the label and the data even the label is randomized. Their experiment shows that the time of overfitting of their models increases with the error label rate. In other words, the network needs more time to learn the noisy dataset. Based on this phenomenon, there are two assumptions based on our intuition.

\begin{enumerate}
    \item The label corruption makes the relationship between the data and the label more complex. It makes the cost to describe the relationship increases, which means the network can accumulate more information in training.
    \item The label corruption makes the dataset contains more conflict information, which decreases the quantity of information accumulated by the network.
\end{enumerate}
The performance-based evaluation is not useful to answer these questions, and we use our method to verify these two assumptions. Briefly, if the first assumption is correct, the network accumulates more information in the same length of time. Otherwise, if the second assumption is correct, the network accumulates less information within the same time.

We use TensorFlow CNN Demo as the test model and CIFAR-10 as the test dataset with the same training configuration shown in Tab. \ref{Tab_traininginfo}. The experiment process is shown in Exp. \ref{labelcorruption_exp}.

\begin{algorithm}[ht]
\caption{Impact of label corruption \label{labelcorruption_exp}}
\begin{algorithmic}[1]
    \STATE Create eleven datasets based on CIFAR-10, $\{A_i\} (i\in [0,100])$. In $A_i$, $(i\times 10)\%$ labels are shuffled randomly.
    \STATE Initialize 10 networks by the same initial weights.
    \STATE Use these datasets to train the model separately within the same time.
    \STATE calculate the distance $d(w_i, w)$, where $w_i$ is the current weights, and $w$ is the final weights after training.
\end{algorithmic}
\end{algorithm}

\begin{figure}[htb]
    \centering
    \includegraphics[width=0.44\textwidth]{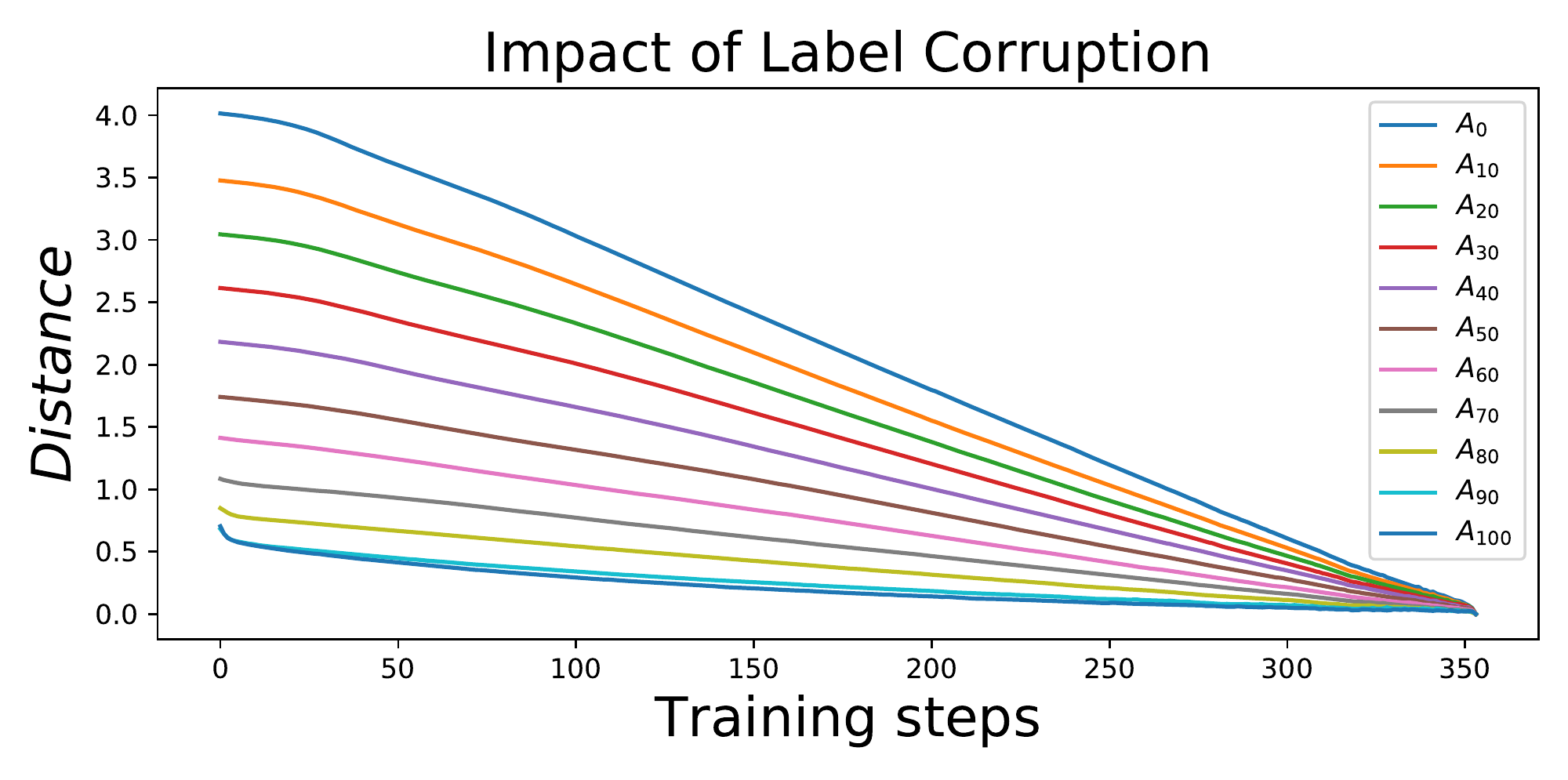}
    \caption{The label corruption can decrease the quantity of the information provided by the dataset.}
    \label{Label_corruption}
\end{figure}

The result is shown in Fig. \ref{Label_corruption}. The intercept of the curve shows that the amount of the information accumulated by the network with the label error rate increasing, which means the label corruption hinders the network's learning.

\section{Discussion}
\textbf{Question about information effectiveness}. This paper provides a tool to analyze the information accumulated by the network in the training process. The precondition of this method is that most of the information learned by the network needs to be useful, which means the training process's output needs to be fine-tuned. Otherwise, we cannot guarantee that the nearest points of the initial weights in $supp(P)$ might not be the output weights.

However, in practice, this precondition is not always satisfied. The best counterexample is the overfitting phenomenon, which shows that the network can perform well on the training dataset but performs badly on the test dataset. One of the accepted explanations is that the network learned too much knowledge from the training dataset, not the commonality between the training dataset and the test dataset. It indicates that not all the information learned by the network is useful and meaningful.

Therefore, to analyze such failed training cases, we still need further research to quantify the information's effectiveness.

\textbf{Question about cross-modal verification}. We use four datasets from simple to complex (MNIST, CIFAR-10, TensorFlow Flowers, Pascal VOC). And we use seven models to accept the information from these four from simple to complex (TF MNIST Demo, TF CNN Demo, AlexNet, VGG, ResNet, GoogleNet, Yolo). Although our method is verified by all of these models when the model's structure is fixed, the size order of the information quantity is not consistent with our expectations when the network structure is different (see Tab. \ref{Tab_distanceorder}).

On the one hand, to train the model to fine-tuned, we use different training configuration to train the model as Tab. \ref{Tab_traininginfo} shows, which might impact the information accumulation. On the other hand, for different models, their capacity of representation is different. It means to store a specific piece of information, the bigger ones' weights need less change than the smaller ones', which means the presented information quantity is less than the others.
Even though we do not deny that this reflects our research's limitations, it indicates that we need further study to provide a more general model based on the current achievement.


\begin{table}[]
    \centering
    \scalebox{0.85}{
    \begin{tabular}{|c|c|c|}
    \hline
    Rank & Expected order &  Actual order  \\
    \hline
    1& TF MNIST Demo (MNIST) & VGG (TF Flower) \\
    2& TF CNN Demo (cifar-10) & TF MNIST Demo (MNIST)\\
    3& AlexNet (cifar-10) & TF CNN Demo (cifar-10)\\
    4& VGG (TF Flower)& AlexNet (cifar-10)\\
    5& ResNet (TF Flower)& Yolo (Pascal VOC)\\
    6& GoogleNet (TF Flower) & GoogleNet  (TF Flower)\\
    7& Yolo v3 (Pascal VOC) & ResNet  (TF Flower)\\
    \hline
    \end{tabular}}
    \caption{The information quantity rank in the experiments.}
    \label{Tab_distanceorder}
\end{table}

\section{Future Works}
Even though this work still gives us a new view to analyze the essence of the neural network. Quantify the information is great progress to answer the questions about the network explanation. We will apply it in the following fields to solve the related questions.
\begin{enumerate}
    \item By quantifying the information in a different part of the network, we can target the data's key feature with more confidence.
    \item By quantifying the network's information, we can evaluate the training process and optimize it.
    \item By analyzing the information quantity changing, we can reveal the essence of the double decent phenomenon \cite{nakkiran2019deep}.
\end{enumerate}
Moreover, we will keep working in this field for a more general model.

\bibstyle{aaai21}

\end{document}